\documentclass[journal]{IEEEtran}
\usepackage[utf8]{inputenc}
\usepackage[T1]{fontenc}
\usepackage[nocompress]{cite}
\usepackage{amsmath,amssymb}
\usepackage{graphicx}
\usepackage{booktabs}
\usepackage{tabularx}
\usepackage{array}
\usepackage{url}
\usepackage[pdfborder={0 0 1},linkbordercolor={1 0 0},citebordercolor={0 1 0},urlbordercolor={0 1 1}]{hyperref}

\hypersetup{hidelinks}

\newcommand{\covStrong}{\ensuremath{\bullet}}
\newcommand{\covModerate}{\ensuremath{\odot}}
\newcommand{\covWeak}{\ensuremath{\circ}}

\begin{document}

\title{Towards Autonomous Formulaic Alpha Discovery: An Evolutionary Computation Perspective}

\author{Xinwei~Yu,
Yiyang~Fu,
Mingcheng~Fan,
Enqi~Li,
Yilin~Gao,
and~Shugong~Xu,~\IEEEmembership{Fellow,~IEEE}%
\thanks{Yiyang Fu, Enqi Li, and Yilin Gao are with Shanghai University,
Shanghai 200444, China.}%
\thanks{Xinwei Yu, Mingcheng Fan, and Shugong Xu are with Xi'an Jiaotong-Liverpool University,
Suzhou 215123, China.}%
\thanks{Corresponding author: Shugong Xu (e-mail: Shugong.Xu@xjtlu.edu.cn).}}

\maketitle

\begin{abstract}
Automated formulaic alpha discovery aims to generate predictive and interpretable trading signals from large symbolic factor spaces. Its effectiveness is constrained by noisy fitness estimates, market nonstationarity, costly backtesting, semantic redundancy, and conflicting practical objectives. Existing studies employ diverse techniques, including genetic programming (GP), evolutionary algorithms (EAs), reinforcement learning (RL), generative flow networks (GFlowNets), Monte Carlo tree search (MCTS), large language models (LLMs), and agentic workflows, but generally examine them as separate algorithmic families. This article introduces, for the first time, a unified evolutionary computation (EC) perspective on automated formulaic alpha discovery, formulating it as a noisy, dynamic, and multiobjective symbolic evolutionary optimization problem. A six-component analytical framework is developed to characterize existing methods through representation, variation, fitness evaluation, selection, memory, and adaptation. Furthermore, an eight-dimensional, autonomy-oriented evaluation framework is proposed, covering search efficiency, fitness reliability, residual alpha quality, economic diversity, tradability, evolutionary autonomy, robustness to nonstationarity, and reproducibility. Together, these frameworks provide a systematic foundation for unifying heterogeneous approaches, diagnosing component-level limitations, and guiding the development of reliable, adaptive, interpretable, and reproducible autonomous alpha discovery systems.
\end{abstract}


\section{Introduction}
\label{introduction}

Automated formulaic alpha discovery aims to generate predictive and interpretable
trading signals from large symbolic factor spaces. In quantitative
investment, such signals are expected not only to forecast cross-sectional
equity returns, but also to support inspection, risk diagnosis,
neutralization, deployment, monitoring, and reuse in factor libraries
\cite{kakushadze_2016,grinold_and_kahn_2000,brock_et_al_1992,lo_et_al_2000,fama_and_french_1993}.
This requirement distinguishes alpha discovery from conventional
financial prediction \cite{gu_et_al_2020}. A useful alpha should have a
transparent structure that can be evaluated, combined, and governed
within a quantitative research pipeline, rather than only achieving high
historical accuracy.

Early alpha construction mainly relied on human hypotheses, economic
intuition, and manually designed formula libraries
\cite{jegadeesh_and_titman_1993,carhart_1997,asness_et_al_2013}.
These libraries provide reusable benchmarks, but their scalability is
limited in modern financial markets. The number of possible
transformations over price, volume, order flow, fundamental, and
alternative data grows combinatorially, whereas manual hypothesis
generation remains slow and costly. Moreover, effective signals may decay
as market regimes shift, trading crowds form, liquidity conditions
change, and implementation costs increase. These limitations have driven
the transition from human-designed factors to automated formulaic alpha discovery.

\begin{figure}[!t]
\centering
\includegraphics[width=\columnwidth]{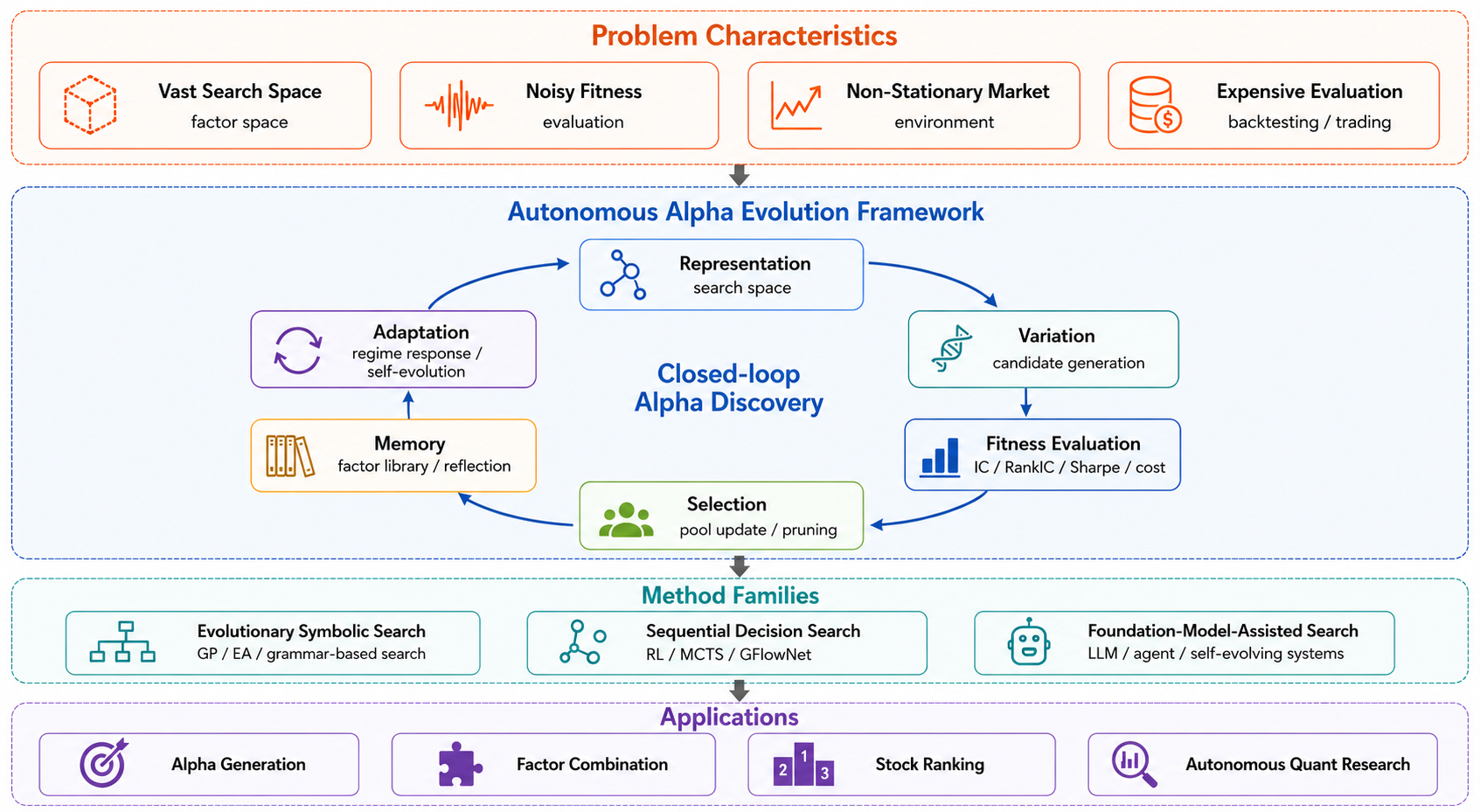}
\caption{A general framework for automated formulaic alpha discovery from an autonomous evolutionary perspective.}
\label{fig:framework}
\end{figure}

Automated formulaic alpha discovery is naturally connected to evolutionary
computation (EC). The search is conducted over a vast symbolic factor
space, guided by noisy empirical fitness, affected by nonstationary
markets, and constrained by costly backtesting and trading evaluation.
Candidate alphas are commonly assessed by metrics such as the information
coefficient (IC), rank information coefficient (RankIC), long--short
return, turnover, Sharpe ratio, and transaction-cost-adjusted return
\cite{grinold_and_kahn_2000}. However, these metrics are noisy estimates
of future utility rather than fixed objective values. These properties make automated formulaic alpha discovery a noisy, dynamic symbolic evolutionary optimization problem involving redundant expressions and multiple practical objectives. 

Recent studies have introduced increasingly automated mechanisms for
generating, refining, and selecting alpha candidates. Population-based
search methods, including genetic programming (GP), search based on
evolutionary algorithms (EAs), and grammar-constrained search, explore
symbolic spaces through variation and fitness-driven selection
\cite{zhang_autoalpha_2020,cui_et_al_2021}. Sequential and distributional
generation methods, including reinforcement learning (RL), Monte Carlo
tree search (MCTS), and Generative Flow Networks (GFlowNets), formulate
candidate construction as policy learning, tree exploration, or
reward-proportional sampling
\cite{yu_et_al_2023,ren_riskminer_2024,chen_alphasage_2025,shi_alpha_jungle_2026}.
More recent foundation-model-guided and agentic systems introduce large
language model (LLM) priors, tool use, memory, reflection, and iterative
research workflows, moving the field toward autonomous quantitative
research systems
\cite{tang_alphaagent_2025,wang_factorminer_2026}.

Despite this progress, existing studies are often reviewed according to
algorithmic labels. Such a view is useful for tracing technical lineage,
but it may obscure the common evolutionary structure shared by these
methods. GP, RL, GFlowNets, MCTS, LLM-guided generation, and agentic
systems are not independent conceptual dimensions. Rather, they automate
different parts of a common discovery loop, including representation,
variation, fitness evaluation, selection, memory, and adaptation. The central argument is that automated formulaic alpha discovery should
be understood not as a sequence of isolated methodological waves, but as
noisy, dynamic symbolic evolutionary optimization.

Fig.~\ref{fig:framework} summarizes the proposed perspective. The upper
part identifies the main problem characteristics of automated formulaic alpha
discovery, including a vast search space, noisy fitness, nonstationary
markets, and costly evaluation. The middle part organizes the
discovery process as a closed evolutionary loop with six components:
representation, variation, fitness evaluation, selection, memory, and
adaptation. The lower part connects this loop to representative method
families and applications, including alpha generation, factor
combination, stock ranking, and autonomous quantitative research. This
overview emphasizes that increasing automation is meaningful only when
the complete discovery loop is considered.

The main contributions are fourfold.
\begin{enumerate}
\item \emph{EC-oriented problem reframing.} Automated formulaic alpha
discovery is formulated as noisy, dynamic symbolic evolutionary optimization.

\item \emph{Six-component analytical framework.} Representative methods are
compared through representation, variation, fitness evaluation, selection,
memory, and adaptation.

\item \emph{Component-level taxonomy.} Existing studies are organized by method
family, evidence level, component coverage, evaluation setting,
reproducibility evidence, and limitation.

\item \emph{Autonomy-oriented evaluation roadmap.} A reliability-centered
evaluation protocol is proposed for future autonomous alpha discovery systems.
\end{enumerate}

The remainder of this article is organized as follows.
Section~\ref{background-formulaic-alpha-discovery} defines automated formulaic alpha discovery, characterizes its symbolic search space, and relates its
core difficulties to EC.
Section~\ref{unified-framework-for-autonomous-alpha-evolution} presents
the six-component autonomous evolutionary framework.
Section~\ref{taxonomy-of-automated-alpha-discovery-methods} develops the
component-level taxonomy and reviews method families from human-guided
formula construction to foundation-model-guided and agentic discovery.
Section~\ref{evaluation-protocol-and-future-roadmap} formulates the
autonomy-oriented evaluation protocol and discusses benchmarks, reporting,
factor governance, and roadmap issues.
Section~\ref{conclusion} concludes this article.

\section{Background: Formulaic Alpha Discovery}
\label{background-formulaic-alpha-discovery}

This section provides the theoretical background for automated formulaic
alpha discovery. It first defines the search target, characterizes the
symbolic search space, and treats empirical financial metrics as noisy
fitness observations rather than fixed objective values. It then explains
why evolutionary computation (EC) provides a natural methodological
foundation for the six-component autonomous evolutionary framework for
automated formulaic alpha discovery. This background links the motivation presented in Section~\ref{introduction} to the six-component autonomous evolutionary framework introduced in Section~\ref{unified-framework-for-autonomous-alpha-evolution}.

\subsection{Formulaic Alpha Factors}
\label{formulaic-alpha-factors}

A \emph{formulaic alpha factor} is an interpretable symbolic expression
that maps primitive market observations to cross-sectional predictive
scores. Given a set of primitive market fields, symbolic operators, and
temporal or cross-sectional windows, a candidate alpha can be expressed as
\begin{equation}
    \alpha = g(\mathcal{X}, \mathcal{O}, \mathcal{W}),
    \label{eq:formulaic-alpha}
\end{equation}
where $\alpha$ denotes a candidate formulaic alpha expression,
$\mathcal{X}$ denotes primitive market fields, such as price,
volume, turnover, order flow, fundamental, or alternative data;
$\mathcal{O}$ denotes symbolic operators; $\mathcal{W}$ denotes temporal
or cross-sectional windows; and $g(\cdot)$ denotes a compositional mapping
from raw observations to a symbolic alpha expression.

For a stock universe $\mathcal{U}_{t}$ at time $t$, the output of a
formulaic alpha is a cross-sectional score vector
\begin{equation}
    \mathbf{z}^{\alpha}_{t}
    =
    \left[
    z^{\alpha}_{1,t}, z^{\alpha}_{2,t}, \ldots,
    z^{\alpha}_{|\mathcal{U}_{t}|,t}
    \right]^{\top},
    \label{eq:alpha-score-vector}
\end{equation}
where $z^{\alpha}_{i,t}$ denotes the score assigned to stock $i$
by formula $\alpha$ at time $t$.
These scores are typically used for stock ranking, portfolio
construction, risk diagnosis, or factor-library management. In contrast
to black-box return predictors, formulaic alphas preserve explicit
symbolic structures, thereby facilitating inspection, neutralization,
combination, monitoring, and reuse.

Before automated discovery, formulaic alphas were mainly constructed by
human experts using economic hypotheses and manually designed formula
libraries
\cite{kakushadze_2016,grinold_and_kahn_2000,brock_et_al_1992,lo_et_al_2000,fama_and_french_1993}.
Such libraries encode important financial intuition and provide reusable
benchmarks. However, their scalability is limited because the number of
possible transformations over primitive market fields increases
combinatorially. Automated formulaic alpha discovery extends this process
from manual hypothesis construction to machine-driven symbolic search, in
which candidate formulas are generated, evaluated, selected, stored, and
adapted under changing market conditions.

\subsection{Symbolic Search Space}
\label{symbolic-search-space}

The search space of formulaic alpha discovery is a compositional symbolic
space. A candidate expression is constructed from a domain-specific
language (DSL) that specifies primitive fields, operators, constants,
window parameters, and syntactic rules. A typical DSL may contain
primitive fields such as $\mathrm{open}$, $\mathrm{high}$, $\mathrm{low}$,
$\mathrm{close}$, $\mathrm{volume}$, $\mathrm{vwap}$, $\mathrm{amount}$,
and order flow variables; unary operators such as $\log(\cdot)$,
$\mathrm{abs}(\cdot)$, $\mathrm{sign}(\cdot)$, and
$\mathrm{rank}(\cdot)$; binary operators such as $+$, $-$, $\times$, and
$/$; time-series operators such as $\mathrm{ts_mean}(\cdot)$,
$\mathrm{ts_std}(\cdot)$, $\mathrm{ts_corr}(\cdot)$, and
$\mathrm{ts_rank}(\cdot)$; and cross-sectional transformations such as
$\mathrm{rank}(\cdot)$, neutralization, winsorization, and
standardization.

Formally, let $\Omega$ denote the set of all syntactically valid
expressions induced by a DSL. Automated formulaic alpha discovery searches for candidate expressions
\begin{equation}
    \alpha \in \Omega.
    \label{eq:alpha-search-space}
\end{equation}
The difficulty of this search does not arise solely from the size of
$\Omega$, but also from its structural properties. First, the space is
combinatorial because primitive fields, operators, constants, windows, and
cross-sectional transformations can be recursively composed. Second, the
space is constrained because expressions must satisfy type consistency,
operator arity, window validity, and grammar legality. Third, the space
is redundant because syntactically distinct expressions may produce
similar factor values, exposures, or return streams. Fourth, the space is
sparse because most valid expressions are economically implausible,
statistically unstable, redundant, or nontradable.

Different methods impose different representations on this symbolic
space. Genetic programming commonly represents candidate formulas as
expression trees. Grammar-based search represents candidates as
derivations from a formal grammar. RL-based methods represent formula construction as a sequential decision process. MCTS explores partial expressions through a search tree. GFlowNet-based methods aim to
sample diverse high-reward expressions. LLM-guided and agent-based methods may represent formulas as textual programs, tool-executable expressions, or memory-augmented research artifacts. These representations differ in implementation, but they all address the same fundamental problem: how to efficiently search a vast, constrained, redundant, and sparse symbolic expression space
\cite{schmidt_and_lipson_2009,vladislavleva_et_al_2009,udrescu_and_tegmark_2020,la_cava_et_al_2021}.

\subsection{Empirical Fitness as a Noisy Objective}
\label{empirical-fitness-signals}
In EC, fitness determines the selection pressure that guides search. In
formulaic alpha discovery, however, fitness is not directly observed. It
is estimated from empirical financial data through backtesting,
cross-sectional prediction, portfolio simulation, or risk-adjusted
performance evaluation. These metrics are not fixed objective values;
they are noisy empirical estimates of future utility. They guide
variation and selection, but they are affected by finite samples,
temporal dependence, repeated testing, data leakage, regime shifts, and
implementation assumptions.

Common empirical proxies include IC, RankIC, ICIR, long--short return,
turnover, and transaction-cost-adjusted return. The information
coefficient (IC) measures cross-sectional predictive association:
\begin{equation}
    \mathrm{IC}_{t}
    =
    \mathrm{corr}\left(\mathbf{z}^{\alpha}_{t}, \mathbf{r}_{t+1}\right),
    \label{eq:ic}
\end{equation}
where $\mathbf{z}^{\alpha}_{t}$ denotes the score vector generated by
candidate formula $\alpha$ and $\mathbf{r}_{t+1}$ denotes future returns.

RankIC replaces linear correlation with rank correlation:
\begin{equation}
    \mathrm{RankIC}_{t}
    =
    \mathrm{corr}_{S}\left(\mathbf{z}^{\alpha}_{t}, \mathbf{r}_{t+1}\right),
    \label{eq:rankic}
\end{equation}
where $\mathrm{corr}_{S}(\cdot,\cdot)$ denotes the Spearman rank
correlation. ICIR summarizes the temporal stability of IC:
\begin{equation}
    \mathrm{ICIR}
    =
    \frac{\mathbb{E}_{t}[\mathrm{IC}_{t}]}
    {\sigma_{t}(\mathrm{IC}_{t})},
    \label{eq:icir}
\end{equation}
where $\mathbb{E}_{t}[\mathrm{IC}_{t}]$ and
$\sigma_{t}(\mathrm{IC}_{t})$ denote the time-series mean and standard
deviation of IC.

Portfolio-based proxies connect predictive ranking to economic payoff.
The long--short return is
\begin{equation}
    R^{\mathrm{LS}}_{t}
    =
    \frac{1}{|Q^{\mathrm{top}}_{t}|}
    \sum_{i \in Q^{\mathrm{top}}_{t}} r_{i,t+1}
    -
    \frac{1}{|Q^{\mathrm{bottom}}_{t}|}
    \sum_{i \in Q^{\mathrm{bottom}}_{t}} r_{i,t+1},
    \label{eq:long-short-return}
\end{equation}
where $Q^{\mathrm{top}}_{t}$ and $Q^{\mathrm{bottom}}_{t}$ denote the
top-ranked and bottom-ranked stock groups. Turnover measures the trading
intensity induced by a factor-based portfolio:
\begin{equation}
    \mathrm{Turnover}_{t}
    =
    \frac{1}{2}
    \sum_{i \in \mathcal{U}_{t}}
    \left|
    w_{i,t} - w_{i,t-1}
    \right|,
    \label{eq:turnover}
\end{equation}
where $w_{i,t}$ denotes the portfolio weight of stock $i$. The
transaction-cost-adjusted return is
\begin{equation}
    R^{\mathrm{net}}_{t}
    =
    R^{\mathrm{gross}}_{t}
    -
    c \cdot \mathrm{Turnover}_{t},
    \label{eq:cost-adjusted-return}
\end{equation}
where $R^{\mathrm{gross}}_{t}$ denotes gross return and $c$ denotes the
unit transaction-cost rate.

The role of these metrics is not to provide a financial evaluation tutorial. Rather, they show why alpha discovery is noisy
empirical fitness optimization. A high in-sample IC, RankIC, ICIR, or
net return may reflect genuine predictive structure, but it may also
arise from sampling noise, repeated testing, leakage, or regime-specific
artifacts. Formulaic alpha discovery should be evaluated as a
multiobjective and noisy fitness-estimation problem rather than as the
maximization of a single deterministic score.

\subsection{Evolutionary Computation View of Automated Formulaic Alpha Discovery}
\label{evolutionary-computation-view-of-alpha-discovery}

Formulaic alpha discovery is naturally connected to EC because its core
discovery process can be decomposed into representation, variation,
fitness evaluation, selection, memory, and adaptation
\cite{holland_1975,goldberg_1989}. Candidate formulas are represented as
symbolic genotypes. Their realized factor values, rankings, exposures,
and return streams constitute behavioral phenotypes. Fitness signals are
estimated through empirical financial evaluation. Variation operators
generate new candidates. Selection mechanisms retain promising formulas.
Memory stores validated factors or useful search experience. Adaptation
updates the search process under market nonstationarity.

This connection can be formalized as the following evolutionary loop:
\begin{equation}
    \mathcal{C}_{t+1}
    =
    \operatorname{Sel}
    \left(
    \mathcal{C}_{t}
    \cup
    \mathcal{V}(\mathcal{C}_{t}, \mathcal{M}_{t}),
    \hat{F}_{t}
    \right),
    \label{eq:evolutionary-loop}
\end{equation}
where $\mathcal{C}_{t}$ denotes the active candidate set or population at
search iteration $t$, $\mathcal{V}(\cdot)$ denotes a variation or generation
operator, $\mathcal{M}_{t}$ denotes memory, $\hat{F}_{t}$ denotes the noisy
empirical fitness estimator, and $\operatorname{Sel}(\cdot)$ denotes the
selection operator. The retained validated factor pool is denoted by
$\mathcal{B}_{t}$. Under this view, automated formulaic alpha discovery is better understood as a closed-loop optimization process rather than a one-shot procedure for generating formulas, since each search step is shaped by uncertain and time-varying feedback.

\subsubsection{Noisy Fitness Optimization}

The empirical fitness signal is noisy because each candidate is evaluated from
finite samples, correlated cross-sectional observations, repeated tests,
and backtesting assumptions. The observed fitness of a candidate factor
can be written as
\begin{equation}
    \hat{F}_{t}(\alpha)
    =
    F^{\ast}_{t}(\alpha)
    +
    \epsilon_{t}(\alpha),
\end{equation}
where $\hat{F}_{t}(\alpha)$ denotes the observed empirical fitness,
$F^{\ast}_{t}(\alpha)$ denotes the latent predictive utility, and
$\epsilon_{t}(\alpha)$ denotes estimation noise. This noise can mislead
selection because candidates with inflated in-sample scores may be
selected even when their true predictive utility is weak. For this reason, alpha discovery depends not only on the choice of fitness metric, but also on fitness reliability, control of repeated testing, validation design, and out-of-sample robustness \cite{campbell_and_thompson_2008,goyal_and_welch_2008}.

\subsubsection{Dynamic Evolutionary Optimization}

Alpha discovery is also a dynamic optimization problem because market conditions change over time. The optimal expression is not fixed, but depends on the current market environment:
\begin{equation}
    \alpha^{\ast}_{t}
    =
    \arg\max_{\alpha \in \Omega}
    F_{t}(\alpha),
    \label{eq:dynamic-alpha}
\end{equation}
where $F_{t}(\alpha)$ denotes the time-varying fitness landscape. Market regime shifts, investor crowding, liquidity changes, and factor decay can all
move the optimum. This implies that a system optimized only for
historical fitness may fail when the market environment changes.
Adaptation is a necessary component of autonomous alpha evolution
\cite{angeline_1997,morrison_2004,mclean_and_pontiff_2016}.

\subsubsection{Quality-Diversity Search}

Alpha discovery should not converge to a single best formula. In
quantitative investment, a useful system should construct a diverse
factor library in which factors provide complementary signals. This
requirement connects alpha discovery to quality-diversity search. A
validated factor archive can be described as
\begin{equation}
    \mathcal{B}_{t}
    =
    \left\{
    \alpha_k \in \mathcal{C}_{t}
    \mid
    F_{t}(\alpha_k) \geq \tau,\;
    d_{t}(\alpha_k,\alpha_l) \geq \delta,\;
    k \neq l
    \right\}.
    \label{eq:alpha-archive}
\end{equation}
where $\mathcal{B}_{t}$ denotes the retained factor pool at iteration $t$,
$\tau$ denotes a minimum quality threshold, $d_{t}(\cdot,\cdot)$ denotes a
behavioral distance measure, and $\delta$ denotes a minimum diversity
threshold. This formulation emphasizes that diversity should be evaluated behaviorally, for example, through factor correlation, exposure similarity, return-stream correlation, turnover patterns, or residual alpha contribution.

\subsubsection{Multiobjective Optimization}

A practically useful alpha should be predictive, stable, simple, diverse,
robust, tradable, and reproducible. These requirements often conflict.
For this reason, alpha discovery is more naturally formulated as a
multiobjective optimization problem
\cite{coello_coello_et_al_2007,zitzler_and_thiele_1999,knowles_and_corne_2000}:
\begin{equation}
    \begin{split}
    \max_{\alpha \in \Omega}
    \mathbf{F}_{t}(\alpha)
    =
    \left[
    F_{\mathrm{pred},t}(\alpha),\;
    F_{\mathrm{stab},t}(\alpha),\;
    F_{\mathrm{div},t}(\alpha),\right.\\
    F_{\mathrm{trade},t}(\alpha),\;
    F_{\mathrm{simp},t}(\alpha),\;
    F_{\mathrm{repr},t}(\alpha)]^{\top},
    \end{split}
    \label{eq:multiobjective-alpha}
\end{equation}
where $F_{\mathrm{pred}}$, $F_{\mathrm{stab}}$, $F_{\mathrm{div}}$,
$F_{\mathrm{trade}}$, $F_{\mathrm{simp}}$, and $F_{\mathrm{repr}}$ denote
predictive quality, stability, diversity and novelty, tradability,
simplicity and interpretability, and reproducibility, respectively.
Each component is observed through noisy empirical proxies rather than
directly measured utility. This formulation shows why a single scalar fitness value is often insufficient and why evaluation protocols should report several complementary dimensions.

\subsubsection{Semantic Redundancy Control}

Symbolic search spaces often contain substantial redundancy. Formulaic expressions may differ in syntax while producing similar behavior. Let $\phi(\alpha)$ denote the behavioral representation of a formulaic alpha factor, such as its standardized factor matrix, exposure vector, ranking sequence, or return stream. The semantic similarity between two formulas can then be measured as
\begin{equation}
    \rho_{kl}
    =
    \mathrm{corr}
    \left(
    \phi(\alpha_k), \phi(\alpha_l)
    \right),
    \label{eq:semantic-similarity}
\end{equation}
where $\rho_{kl}$ denotes the behavioral similarity between candidates
$\alpha_k$ and $\alpha_l$. Redundancy control aims to prevent the archive from being filled with output-equivalent formulas. This is important because a factor library with many correlated alphas may appear large while adding little incremental portfolio value.

\subsection{From Static Formula Mining to Autonomous Alpha Evolution}
\label{from-static-formula-mining-to-autonomous-alpha-evolution}

The preceding analysis shows that formulaic alpha discovery is not merely
a task of generating more formulas. It is a noisy, nonstationary, costly,
redundant, and multiobjective symbolic evolutionary optimization problem.
This characterization motivates a transition from static formula mining
to autonomous alpha evolution.

Static formula mining focuses on whether a search algorithm can generate
candidate expressions with high empirical scores. Autonomous alpha
evolution focuses on whether a system can continuously represent,
generate, evaluate, select, store, reuse, and adapt factors under
nonstationary market conditions. From this perspective, different method
families are not isolated labels, but different implementations of the
same evolutionary loop. GP and evolutionary algorithms (EAs) emphasize
population-based symbolic variation and fitness-driven selection. RL and
MCTS strengthen sequential construction and exploration control.
GFlowNets emphasize diverse reward-proportional generation. LLM-guided
search introduces language priors and tool use. Agentic systems
extend memory, reflection, and adaptation.

This EC-theoretic perspective provides the foundation for the
six-component framework developed in the next section. Representation
defines the genotype space of alpha formulas. Variation defines how new
candidate expressions are generated. Fitness evaluation estimates
predictive and economic utility. Selection determines which candidates
survive. Memory stores validated factors and search experience.
Adaptation updates the system under changing market regimes. Together,
these six components form the autonomous evolutionary view of formulaic
alpha discovery.

\begin{table*}[!t]
\caption{Qualitative Overview of Representative Method Families.}
\label{tab:six-component-method-map}
\centering
\scriptsize
\setlength{\tabcolsep}{3.2pt}
\renewcommand{\arraystretch}{1.6}
\newcolumntype{Y}{>{\raggedright\arraybackslash}X}
\begin{tabularx}{\textwidth}{@{}
>{\raggedright\arraybackslash}p{0.18\textwidth}
Y
Y
Y@{}}
\toprule
Method Family & Search Logic & Main Strength & Main Limitation \\
\midrule

Human-guided formula libraries &
Manual design of symbolic factor templates from economic hypotheses &
Interpretable priors and reusable benchmark formulas &
Limited scalability and strong dependence on human expertise \\

GP- and EA-based symbolic search &
Population-based variation and fitness-driven filtering over symbolic expressions &
Direct instantiation of evolutionary search in formula spaces &
Fitness noise, bloat, and semantic redundancy \\

RL-based sequential generation &
Policy-based construction of formulas through token or operator sequences &
Learned generation policies for large discrete spaces &
Sparse rewards, policy collapse, and fixed historical environments \\

Pool-aware and combination-oriented mining &
Discovery linked to factor-pool construction and downstream combination &
Better alignment with portfolio-level factor contribution &
Candidate pools may inherit first-stage overfitting \\

Diversity-oriented and graph-based search &
Reward-proportional sampling or graph-guided retrieval over candidate factors &
Improved exploration diversity and structured factor relations &
Structural diversity may not imply economic diversity \\

MCTS and grammar-constrained search &
Tree expansion, rollout, and grammar-constrained formula planning &
Budget-aware exploration and explicit search control &
High rollout cost and stationary-reward assumptions \\

LLM-guided formula generation &
Language-prior-driven proposal, repair, and refinement of alpha formulas &
Semantic priors, flexible generation, and tool-assisted iteration &
Prompt sensitivity, hallucination, and dependence on external validation \\

Agentic and self-evolving discovery &
Tool use, memory, reflection, and iterative research workflows &
Toward closed-loop autonomous alpha discovery &
Reproducibility risk, unvalidated memory updates, and meta-overfitting \\

\bottomrule
\end{tabularx}
\end{table*}

\section{Unified Framework for Autonomous Alpha Evolution}
\label{unified-framework-for-autonomous-alpha-evolution}

Throughout this article, automated formulaic alpha discovery denotes the task of
generating interpretable formulaic alpha expressions from symbolic factor
spaces, whereas autonomous alpha evolution denotes the proposed closed-loop
system perspective for analyzing how such expressions are represented,
generated, evaluated, selected, stored, and adapted under financial feedback.
Thus, autonomous alpha evolution is not a separate task, but a framework-level
reframing of automated formulaic alpha discovery.

This section develops the six-component analytical framework used
throughout this article. Building on the EC-theoretic background in
Section~\ref{background-formulaic-alpha-discovery}, automated formulaic
alpha discovery is formulated as a closed-loop evolutionary system that
operates under noisy empirical feedback, nonstationary market
conditions, costly evaluation, semantic redundancy, and
multiobjective constraints. Under this framework, GP, EAs, RL, GFlowNets, MCTS, LLM-guided search, and agentic systems are analyzed as different instantiations of the same autonomous evolutionary discovery process. 

\begin{figure*}[!t]
\centering
\includegraphics[width=0.93\textwidth]{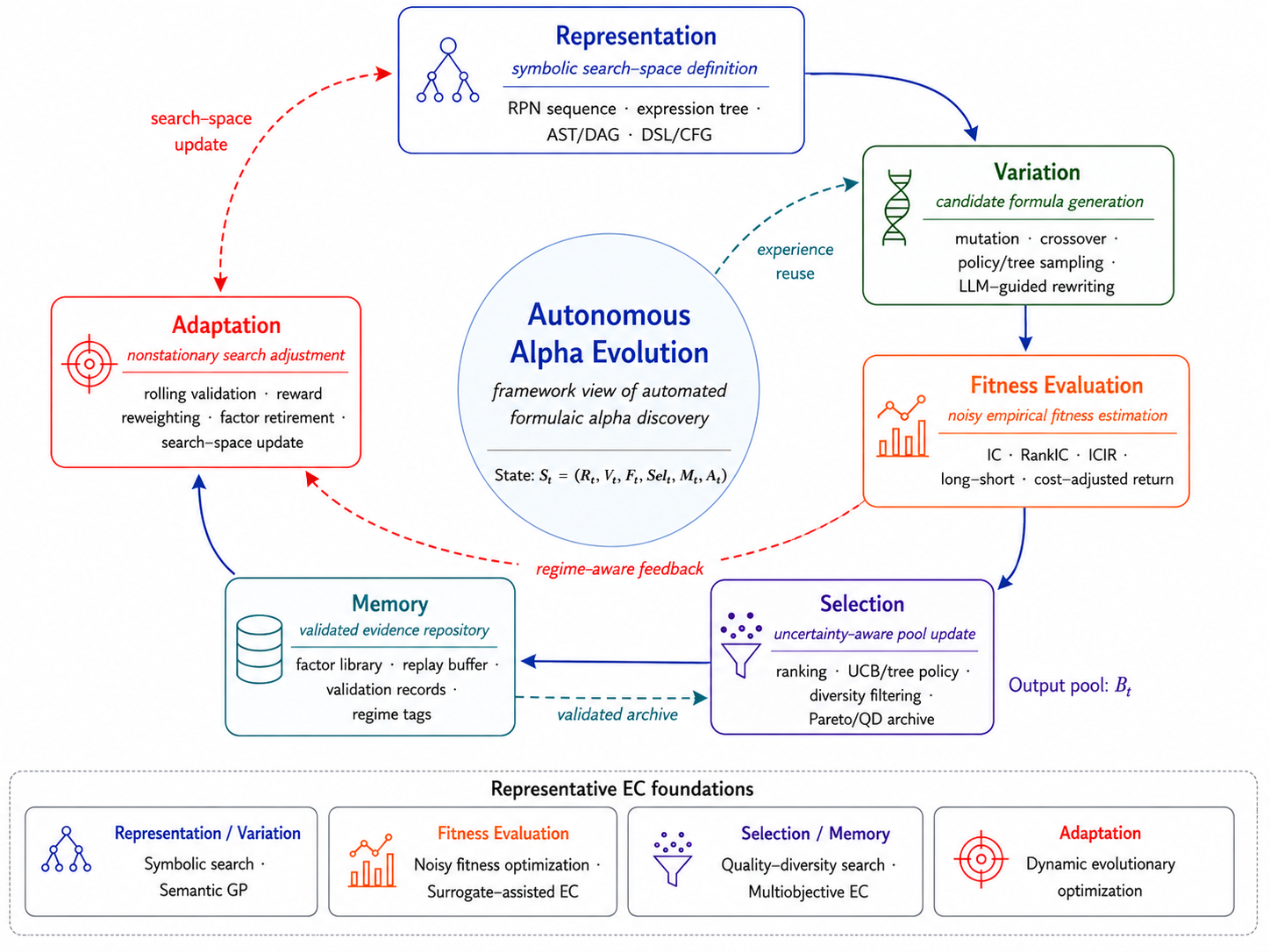}
\caption{Six-component framework for autonomous alpha evolution. The framework organizes automated formulaic alpha discovery as a closed-loop evolutionary system under noisy fitness, nonstationary markets, costly evaluation, and multiobjective financial constraints.}
\label{fig:six-component}
\end{figure*}

\subsection{The Six-Component Framework}
\label{the-six-component-framework}

Automated formulaic alpha discovery can be formulated as an autonomous
evolutionary system that repeatedly represents, generates, evaluates,
selects, stores, and adapts candidate alphas. The analysis shifts from algorithmic labels to functional components. A GP system,
an RL-based generator, a GFlowNet sampler, an MCTS explorer, and an
LLM-based agent system may differ substantially in implementation, but
each must specify how candidate formulas are encoded, how new candidates
are generated, how empirical quality is evaluated, how the retained pool
is updated, how previous experience is stored, and how the search process
is adjusted under changing market conditions.

Fig.~\ref{fig:six-component} illustrates the proposed six-component
framework. The outer environment indicates that alpha discovery is
conducted under noisy, nonstationary, costly, and multiobjective financial
feedback. The six components form a closed loop: representation defines
the symbolic search space; variation generates candidate formulas;
fitness evaluation estimates predictive and economic utility; selection
updates the retained factor pool; memory accumulates validated search
experience; and adaptation adjusts the system in response to regime
changes. The lower part of the figure links the framework to
representative EC foundations, including symbolic search, noisy fitness
optimization, dynamic evolutionary optimization, quality-diversity
search, multiobjective EC, surrogate-assisted EC, and semantic GP
\cite{jin_2011}.

At search iteration $t$, the state of an automated formulaic alpha discovery system can be represented as
\begin{equation}
    \mathcal{S}_{t}
    =
    \left(
    \mathcal{R}_{t},
    \mathcal{V}_{t},
    \mathcal{F}_{t},
    \operatorname{Sel}_{t},
    \mathcal{M}_{t},
    \mathcal{A}_{t}
    \right),
    \label{eq:six-component-state}
\end{equation}
where $\mathcal{R}_{t}$ specifies the representation of the search space, $\mathcal{V}_{t}$ is the variation mechanism used to generate new candidate alphas, $\mathcal{F}_{t}$ is the empirical fitness evaluator, $\operatorname{Sel}_{t}$ is the selection mechanism that filters, ranks, or retains candidates under noisy empirical fitness, $\mathcal{M}_{t}$ stores information accumulated from previous search and evaluation episodes, and $\mathcal{A}_{t}$ is the adaptation mechanism that responds to regime changes and shifts in the fitness landscape. The retained factor pool produced by selection is denoted by $\mathcal{B}_{t}$. Here, $\operatorname{Sel}_{t}$ refers to the decision rule, while $\mathcal{B}_{t}$ refers to the resulting archive or pool. The first four components extend the classical evolutionary loop, whereas memory and adaptation transform static formula search into autonomous discovery.

The six components are coupled rather than independent. Representation
determines which variation operations are valid. Variation determines
which regions of the search space are explored. Fitness evaluation provides the objective signal used by selection. Selection determines
which candidates enter the factor pool and which experiences become
available for future use. Memory influences subsequent generation and
selection by retaining validated formulas, failed trials, or search
trajectories. Adaptation modulates the other components when the market
environment changes. This coupling explains why progress in automated
formulaic alpha discovery cannot be judged only by the sophistication of the generator or by the highest in-sample IC. When the empirical fitness signal is unreliable, more expressive representations, stronger variation operators, and larger memories may amplify historical noise rather than improve discovery.

The component mapping also relates existing method families to EC theory. GP and EAs mainly instantiate symbolic representation, mutation, crossover, and fitness-driven selection. RL methods strengthen the variation component by learning sequential construction policies from previous trials. GFlowNet-based methods emphasize reward-proportional generation and diversity across high-quality expressions. MCTS-based methods make selection and exploration explicit through tree expansion and upper-confidence search. LLM-guided methods inject language priors, operator-level reasoning, and tool use into candidate generation. Agentic systems further introduce structured
memory, reflection, and adaptive research workflows. These developments should not be read as a simple linear replacement of one method by another.
They instead automate different components of the same evolutionary discovery loop.

\subsection{Representation: Defining the Search Space}
\label{representation-defining-the-search-space}

The representation component $\mathcal{R}$ defines the genotype space of
candidate alphas. It determines what expressions can be constructed, how
they are encoded, which syntactic constraints must be satisfied, and how
subsequent components can operate on the encoded candidates. In formulaic
alpha discovery, representation is more than an implementation
choice. It shapes the structure of the search space, the validity of
generated formulas, the interpretability of discovered signals, and the
reproducibility of reported results.

Existing studies have used several representation families. Formulaic GP and
EA-based methods commonly represent alphas as expression trees or
operator compositions, enabling subtree mutation, crossover, and direct
fitness evaluation. Token-based methods use reverse Polish notation (RPN) or other sequence encodings, making formula generation compatible with sequence models and policy learning. Grammar-based methods use domain-specific languages (DSLs) or context-free grammars (CFGs) to ensure syntactic validity and constrain the search to executable expressions. Graph-based approaches represent formulas as abstract syntax trees (ASTs), directed acyclic graphs (DAGs), or factor graphs, which can capture dependency structures and facilitate the reuse of subexpressions. In LLM-guided and agent-based systems, candidates may also be represented as natural-language hypotheses, executable programs, or tool-callable research artifacts.

At the theoretical level, representation mediates the trade-off between searchability and expressiveness. A highly expressive representation may
cover complex alpha structures, but it also expands the search space,
increases the number of invalid candidates, and can intensify semantic redundancy. A highly constrained representation improves validity and
reproducibility, but it may exclude useful financial hypotheses. Representation introduces the first trade-off in the framework:
expressiveness must be balanced against validity, interpretability,
semantic diversity, and evaluation cost.

\subsection{Variation: Generating Candidate Alphas}
\label{variation-generating-candidate-alphas}

The variation component $\mathcal{V}$ determines how new candidate alphas
are generated. In EC, variation drives exploration by moving the population through the search space. In alpha discovery, this component includes formula mutation, crossover, operator substitution, grammar expansion, policy sampling, trajectory generation, tree expansion, and LLM-based rewriting.

The literature shows a clear progression in how variation is implemented.
Classical GP- and EA-based methods rely on random or rule-constrained mutation and crossover to generate new formulas. RL-based methods formulate alpha construction as a sequential decision process and learn policies for
selecting fields, operators, and windows. GFlowNet-based methods aim to
sample diverse high-reward formulas rather than converge to a single
optimization trajectory. MCTS-based methods expand partial formulas in search trees, with explicit control over exploration and exploitation. LLM-guided methods propose, rewrite, or repair formulas using language priors and domain descriptions. Agent-based systems combine generation with tool use, reflection, and iterative refinement.

Although variation has received substantial methodological attention, its value still depends on the reliability of downstream evaluation. A
powerful generator can explore more of the search space, but it may also
identify overfitted patterns more quickly when the fitness signal is
noisy. Variation must produce candidates that are valid, diverse, economically plausible, and evaluable under limited evaluation budgets. 

\subsection{Fitness Evaluation: Estimating Alpha Quality}
\label{fitness-evaluation-estimating-alpha-quality}

The fitness evaluation component $\mathcal{F}$ estimates the quality of
candidate alphas. It is a central component of the framework because selection, memory, and adaptation all depend on its output. In
formulaic alpha discovery, fitness is empirical rather than directly
observable. It is usually estimated through cross-sectional predictive
metrics, portfolio backtests, robustness tests, or
transaction-cost-adjusted evaluation.

Most automated formulaic alpha discovery methods rely on the information
coefficient (IC), rank information coefficient (RankIC), Sharpe ratio,
long--short return, or related backtesting scores. These metrics are
useful, but they remain noisy estimates of latent predictive utility. A
candidate may obtain a high in-sample IC because of repeated testing,
regime-specific effects, data leakage, or sampling variation. A reliable evaluation protocol needs to consider predictive strength, temporal stability, residual value after controlling for known risks, diversity relative to existing factors, formula complexity, turnover, transaction cost, and out-of-sample robustness.

From the EC perspective, the key issue is not only how to define a
reward, but also how to regulate selection pressure under noisy feedback.
If $\mathcal{F}$ places too much weight on raw in-sample IC, the system
may be driven toward backtesting artifacts. If $\mathcal{F}$ incorporates
stability, neutralization, cost adjustment, and diversity, the search is
more likely to retain factors with practical value. This explains why
fitness evaluation is the main bottleneck of automated formulaic alpha discovery. Improvements in representation and variation cannot fully compensate for unreliable fitness evaluation.

\subsection{Selection: Updating the Factor Pool}
\label{selection-updating-the-factor-pool}

The selection component $\operatorname{Sel}_{t}$ determines which candidates survive and enter the retained factor pool. In standard EC, selection increases the frequency of candidates with higher fitness. In alpha discovery, selection plays a broader role because the factor pool must be updated under uncertain fitness, semantic redundancy, and portfolio-level complementarity.

Classical GP and EA systems often use tournament selection, elitism, or
fitness ranking \cite{goldberg_and_deb_1991}. MCTS-based methods
implement selection through tree policies, such as upper-confidence-bound
mechanisms that balance exploitation and exploration. GFlowNet-based
methods use distributional sampling to preserve multiple high-reward
modes. Pool-aware alpha discovery methods further examine whether a candidate
adds incremental value to an existing factor library. Agent-based systems
may implement selection through self-evaluation, debate, critique, or
tool-assisted verification.

Selection in alpha discovery extends beyond ranking. Fitness
differences between formulas are often small relative to estimation
noise, while many high-scoring candidates are semantically redundant.
Therefore, selection should consider both quality and complementarity. A
candidate should be retained only if it is predictive, stable, valid,
sufficiently distinct from existing factors, and useful under realistic
implementation constraints. Selection thus acts as a mechanism for reliability and diversity, rather than as a simple score filter.

\subsection{Memory: Accumulating Validated Experience}
\label{memory-accumulating-validated-experience}

The memory component $\mathcal{M}$ describes how a discovery system stores
and reuses previous experience. Classical evolutionary search often
maintains memory implicitly through the population or an elite archive.
Modern automated formulaic alpha discovery systems use richer memory structures, such as factor libraries, replay buffers, evaluated candidate databases, trajectory histories, validation records, and reasoning traces.

Memory changes how alpha discovery is organized. Without memory, search runs remain largely independent. With memory, the system can reuse validated factors, avoid repeatedly exploring unpromising regions, guide future variation, calibrate surrogate evaluators, and support adaptation to new regimes. GP and EA systems typically rely on populations and elite
archives. RL systems may use replay buffers or policy histories.
GFlowNet-based methods can retain trajectory and reward information.
MCTS stores tree statistics. LLM-based agent systems can store formulas,
evaluation outcomes, critiques, and reflections.

A central risk is memory contamination. If overfitted or spurious factors
are stored as successful experiences, later search episodes may be biased
toward unreliable regions. In a domain where many high-scoring formulas
are false discoveries, memory must be validated before reuse. Useful memory is a curated, evidence-aware repository that records formulas and scores together with evaluation settings, market regimes, robustness evidence, failure cases, and redundancy relationships. 

\subsection{Adaptation: Responding to Market Regime Changes}
\label{adaptation-responding-to-market-regime-changes}

The adaptation component $\mathcal{A}$ accounts for the dynamic nature of
financial markets. In a stationary optimization problem, a search system
can often treat past fitness as informative of future fitness. In alpha
discovery, this assumption is fragile because market regimes, investor
behavior, liquidity conditions, transaction costs, and factor crowding
vary over time. Adaptation is therefore a necessary component of
autonomous alpha evolution.

Adaptation can operate at several levels. At the representation level, the
system may revise the DSL, add new fields, remove invalid operators, or
adjust window ranges. At the variation level, it may reallocate search
resources toward promising regions or increase exploration after regime
shift. At the fitness level, it may change validation windows, update
reward weights, emphasize residual alpha, or increase transaction-cost
penalties. At the selection level, it may retire decayed factors or
increase diversity requirements. At the memory level, it may tag stored
experience by regime, decay outdated evidence, or prioritize recently validated trials \cite{maturana_et_al_2012}.

Most existing methods handle nonstationarity passively through rolling
windows or periodic retraining. Such mechanisms are useful but limited.
Active adaptation requires the system to detect, interpret, or respond to
changes in the fitness landscape. Agent-based and self-evolving systems
have begun to move in this direction through memory, reflection,
feedback-driven revision, and tool-supported monitoring. However,
rigorous adaptation remains underdeveloped, especially when it is
evaluated as an independent component rather than treated as an informal
feature of the system.

\subsection{From Six Components to Method Comparison}
\label{from-six-components-to-method-comparison}

The six-component framework provides a structured basis for comparing
automated formulaic alpha discovery methods. Rather than comparing methods only by algorithmic labels or reported IC values, the comparison examines how each method implements representation, variation, fitness evaluation, selection, memory, and adaptation. The component-level comparison reveals a structural imbalance in the literature. Most methods devote substantial effort to representation and variation, whereas fitness reliability, validated memory, and nonstationary adaptation remain comparatively underdeveloped.

Table~\ref{tab:six-component-method-map} gives a qualitative overview of representative method families, and Table~\ref{tab:family-framework} summarizes their component-level coverage within the six-component framework. Together, these mappings show that algorithmic progress has been uneven. GP and EAs provide a strong foundation for symbolic representation and variation, but often rely on simple empirical fitness and limited adaptation. RL, GFlowNets, and MCTS improve sequential generation, distributional exploration, or explicit search control, yet their performance still depends heavily on the reward signal. LLM-guided and agent-based systems introduce priors, tool use, memory, and reflection, but their validation protocols remain insufficiently standardized. This asymmetry motivates the taxonomy in Section~\ref{taxonomy-of-automated-alpha-discovery-methods} and the evaluation protocol proposed in Section~\ref{evaluation-protocol-and-future-roadmap}.

In summary, the six-component framework organizes a heterogeneous literature into a unified analytical structure. It shows that automated
formulaic alpha discovery is not a sequence of disconnected methodological waves, but a progressive attempt to automate an EC loop under financial
constraints. The central open challenge is to build reliable discovery systems in which representation, variation, fitness evaluation, selection, memory, and adaptation are jointly designed and jointly evaluated.

\section{Taxonomy of Automated Formulaic Alpha Discovery Methods}
\label{taxonomy-of-automated-alpha-discovery-methods}

This section develops a component-level taxonomy of automated formulaic
alpha discovery methods. Following the six-component framework in
Section~\ref{unified-framework-for-autonomous-alpha-evolution}, this
review does not rank algorithmic families by reported performance.
Instead, it examines how each family instantiates representation,
variation, fitness evaluation, selection, memory, and adaptation, and how
the literature has moved from human-designed formula construction toward
increasingly autonomous discovery workflows.

The taxonomy is organized by method family rather than by individual benchmark score. This organization is appropriate because the reviewed studies differ in data sources, stock universes, operators, evaluation horizons, neutralization schemes, transaction-cost assumptions, and reporting standards. A direct numerical comparison across papers would therefore be difficult to interpret and potentially misleading. The purpose of this section is to provide a structured component-level comparison:
which methods exist, which components of the evolutionary loop they
mainly strengthen, what level of evidence supports them, and what
limitations remain visible from the published descriptions.

\begin{table*}[!t]
\caption{Evidence categories used in the taxonomy.}
\label{tab:evidence-categories}
\centering
\scriptsize
\setlength{\tabcolsep}{3pt}
\renewcommand{\arraystretch}{1.35}
\begin{tabularx}{\textwidth}{@{}
>{\raggedright\arraybackslash}p{0.16\textwidth}
>{\raggedright\arraybackslash}p{0.25\textwidth}
>{\raggedright\arraybackslash}X
>{\raggedright\arraybackslash}p{0.22\textwidth}@{}}
\toprule
Evidence Category & Source Type & Role in This Article & Inclusion Treatment \\
\midrule
Peer-reviewed evidence &
Peer-reviewed journal and conference papers &
Serves as primary evidence for method-family analysis, component mapping,
and comparison of established research directions. &
Included as core evidence. \\
\addlinespace[1pt]

Preprint and open-review evidence &
arXiv preprints or OpenReview submissions with complete method descriptions
and experiments &
Captures recent methods that have not yet appeared in archival journals or
conference proceedings. &
Included when the method, data setting, and evaluation protocol are sufficiently
clear. \\
\addlinespace[1pt]

Benchmark and platform evidence &
Benchmark, platform, demonstration, or infrastructure papers &
Supports the analysis of reusable tools, datasets, evaluation platforms,
and engineering settings for alpha discovery. &
Included for taxonomy construction and context; not used alone as conclusive
algorithmic evidence. \\
\addlinespace[1pt]

Technical and repository evidence &
Technical reports, project documents, or repository descriptions &
Provides background information on emerging systems, implementation details,
and industry-facing practice. &
Included only when verifiable method details are available or when used as
background context. \\
\bottomrule
\end{tabularx}
\end{table*}

\subsection{Evidence Organization and Inclusion Scope}
\label{evidence-organization-and-inclusion-scope}

The literature search covers peer-reviewed papers, major conference
proceedings, preprints, benchmark papers, platform papers, and technical
reports related to formulaic alpha discovery. Search sources include
IEEE Xplore, ACM Digital Library, Web of Science, Google Scholar, arXiv,
OpenReview, ACL Anthology, and the proceedings of KDD, AAAI, IJCAI, CIKM,
ICAIF, ICML, ICLR, and NeurIPS
\cite{page_et_al_2021,kitchenham_and_charters_2007,pineau_et_al_2021}.
The search terms cover three groups: formulaic alpha discovery and factor
mining; evolutionary computation, including genetic programming (GP),
symbolic regression, noisy fitness optimization, dynamic optimization,
quality-diversity search, surrogate-assisted evaluation, and semantic GP;
and autonomous discovery, including large language model (LLM) agents,
self-evolving systems, memory-augmented workflows, and tool-based
research systems.

Following established practices for evidence organization and reproducible
research reporting \cite{wilson_et_al_2014,peng_2011}, studies are included
when they satisfy four criteria. First, the search target is a formulaic,
symbolic, programmatic, or otherwise interpretable alpha expression. Second,
the study specifies an automated mechanism for generating, selecting,
retrieving, refining, or combining candidate alpha expressions. Third, the
method is related to evolutionary computation (EC), genetic programming (GP),
evolutionary algorithms (EAs), reinforcement learning (RL), Generative Flow
Networks (GFlowNets), Monte Carlo tree search (MCTS), large language model
(LLM)-guided search, or agent-based discovery. Fourth, the evaluation protocol
is described in sufficient detail to interpret the reported evidence. Purely
neural return predictors without symbolic outputs, portfolio optimization
methods without formula discovery, purely fundamental factor studies, and
materials without verifiable method descriptions are excluded from the main
scope.

Table~\ref{tab:evidence-categories} distinguishes between established and emerging evidence. This distinction matters because recent alpha discovery systems often appear first as preprints, repositories, or technical reports. These materials can reveal important research trends, but their conclusions should remain provisional when peer-reviewed evidence or reproducible evaluation is not yet available.

\subsection{Overview of Method Evolution}
\label{overview-of-method-evolution}

Table~\ref{tab:six-component-method-map} provides a qualitative overview
of representative method families in automated formulaic alpha discovery
from an EC perspective. The table should not be read as a performance
ranking. Instead, it summarizes the dominant search logic, main strength,
and main limitation of each method family, ranging from human-guided
formula libraries to agentic and self-evolving discovery systems. Early
human-designed formula libraries mainly provide representation priors.
GP- and EA-based methods automate symbolic variation and selection.
RL-based methods formulate formula construction as sequential decision
making. Pool-aware and combination-oriented methods connect candidate
generation with retained factor-pool contribution. Diversity-oriented and
graph-based methods, including GFlowNet-based sampling and structured
factor retrieval, strengthen distributional exploration, pool structure,
and memory. MCTS and grammar-constrained methods make exploration control
explicit. LLM-guided methods introduce language priors and reasoning-driven
generation. Agentic and self-evolving systems further operationalize
memory, reflection, tool use, and early forms of adaptation.

The organization in Table~\ref{tab:six-component-method-map} reflects
increasing automation of the discovery loop rather than a guaranteed
improvement in out-of-sample reliability. This distinction is central to this article. More autonomous systems can generate, evaluate, and reuse larger numbers of candidate formulas, but they may also amplify noisy fitness signals
when validation and memory-update rules are weak. The remainder of this section reviews each method family in terms of its main contribution to the six-component framework.

\begin{table*}[!t]
\caption{Literature evolution of automated formulaic alpha discovery.}
\label{tab:literature-evolution}
\centering
\scriptsize
\setlength{\tabcolsep}{2.5pt}
\renewcommand{\arraystretch}{1.30}
\begin{tabularx}{\textwidth}{@{}
>{\raggedright\arraybackslash}p{0.145\textwidth}
>{\raggedright\arraybackslash}p{0.185\textwidth}
>{\raggedright\arraybackslash}X
>{\raggedright\arraybackslash}p{0.215\textwidth}
>{\raggedright\arraybackslash}p{0.155\textwidth}@{}}
\toprule
Stage & Representative Methods & Main Mechanism & Evolutionary Shift & References \\
\midrule
Human-guided formula libraries &
Alpha101, Alpha191, Qlib Alpha158/Alpha360, and classical technical indicators &
Human priors, economic hypotheses, and manually curated interpretable formula libraries &
From discretionary factor design to reusable symbolic factor templates &
\cite{kakushadze_2016,yang_qlib_2020,grinold_and_kahn_2000,gtja_alpha191_2017} \\
\addlinespace[1pt]

GP- and EA-based symbolic search &
AutoAlpha, AlphaEvolve, and GP-based alpha mining systems &
Population-based formula search with mutation, crossover, and fitness-driven filtering &
From manual formula design to automated symbolic search over expression spaces &
\cite{zhang_autoalpha_2020,cui_et_al_2021,koza_1992,eiben_and_smith_2015} \\
\addlinespace[1pt]

RL-based sequential generation &
AlphaGen, Alpha$^{2}$, and AlphaQCM &
Sequential token construction, reward-driven policy learning, and trajectory-level exploration &
From population-level evolution to policy-based sequential formula construction &
\cite{yu_et_al_2023,xu_alpha2_2024,zhu_alphaqcm_2025} \\
\addlinespace[1pt]

Pool-aware and combination-oriented mining &
AlphaForge and related dynamic factor combination methods &
Factor-pool construction, candidate recombination, and downstream portfolio linkage &
From isolated alpha generation to pool-aware discovery and combination-oriented evaluation &
\cite{shi_alphaforge_2025} \\
\addlinespace[1pt]

Diversity-oriented and graph-based search &
AlphaSAGE, AlphaPROBE, and graph-based factor retrieval &
Reward-proportional sampling, factor-pool topology modeling, and structured retrieval &
From single-objective search to diversity-aware exploration and relational factor modeling &
\cite{chen_alphasage_2025,guo_alphaprobe_2026,pugh_et_al_2016} \\
\addlinespace[1pt]

MCTS and grammar-constrained search &
RiskMiner, alphaCFG, and grammar-constrained tree search &
Tree expansion, upper-confidence search, and grammar-constrained formula planning &
From unconstrained generation to budgeted tree search with syntactic and semantic constraints &
\cite{ren_riskminer_2024,shi_alpha_jungle_2026,yang_alphacfg_2026} \\
\addlinespace[1pt]

LLM-guided formula generation &
Alpha-GPT and FAMA &
Natural-language reasoning, symbolic proposal, formula repair, and iterative refinement &
From rule-based generation to language-guided proposal, repair, and refinement &
\cite{wang_alpha_gpt_2025,li_fama_2024} \\
\addlinespace[1pt]

Agentic and self-evolving discovery &
AlphaAgent, FactorMiner, RD-Agent, and QuantEvolver &
Tool use, multiagent coordination, structured memory, reflection, and self-improvement &
From model-guided search to tool-augmented and self-improving discovery workflows &
\cite{tang_alphaagent_2025,wang_factorminer_2026,li_rd_agent_quant_2025,zhang_quant_evolver_2026} \\
\bottomrule
\end{tabularx}
\end{table*}

Table~\ref{tab:literature-evolution} summarizes the historical organization used in this section. The stages are not mutually exclusive. The goal is to capture the dominant methodological shift in each wave. The overall trend is a shift from manually constructed symbolic priors toward systems that automate more components of the EC loop. At the same time, the table shows that later stages mainly increase automation and do not necessarily resolve noisy fitness, redundancy, and nonstationarity.

\subsection{Human-Guided Formula Libraries}
\label{human-guided-formula-construction}

Before automated search, alpha discovery largely followed a human-guided evolutionary process. Researchers proposed market hypotheses, encoded them as symbolic formulas, evaluated them on historical data, and then retained or revised them based on expert judgment \cite{grinold_and_kahn_2000}. The ``101 Formulaic Alphas'' of Kakushadze remain a canonical example: they provide a compact set of symbolic expressions drawn from a much larger implicit search space \cite{kakushadze_2016}. Alpha191, the Alpha158 and Alpha360 collections
in Qlib, and classical technical indicators such as moving averages,
relative strength index, moving average convergence divergence, and
Bollinger Bands serve as related hand-crafted baselines
\cite{yang_qlib_2020,grinold_and_kahn_2000}.

Within the six-component framework, this stage is most closely associated with representation. These formulas define symbolic priors, operator templates, and economic intuitions that later automated systems often inherit.
Variation, selection, memory, and adaptation are mostly carried out by
researchers rather than by an autonomous algorithm. The strength of this
stage lies in interpretability and domain plausibility. Its limitation is
scalability: manual hypothesis evolution cannot cover the combinatorial
expression space, and repeated factor testing may produce false
discoveries when the evaluation protocol is weak
\cite{harvey_et_al_2016}.

\subsection{GP- and EA-Based Symbolic Alpha Search}
\label{gp-and-ea-based-symbolic-alpha-search}

GP- and EA-based methods automate the manual search loop by introducing population-level variation, inheritance, and fitness-driven selection
\cite{koza_1992,eiben_and_smith_2015}. AutoAlpha applies hierarchical GP
with layered search and quality-diversity mechanisms \cite{zhang_autoalpha_2020}. AlphaEvolve reframes alpha generation as an
automated machine learning problem and uses operator design to bias the
search toward parsimonious and relationally informed expressions
\cite{cui_et_al_2021}. These methods form the most direct link between formulaic alpha mining and classical EC.

Within the six-component framework, GP- and EA-based systems mainly strengthen representation, variation, and selection. Candidate alphas are commonly represented as expression trees, formula strings, or symbolic programs. New candidates are generated through mutation, crossover, subtree
replacement, or operator substitution. Selection is usually driven by empirical metrics such as IC, RankIC, returns, or Sharpe ratio.
Population archives provide a limited form of memory, whereas adaptation
to regime changes is usually weak or absent.

The main contribution of this family is the computational template
extended by later systems: generate symbolic expressions, evaluate them empirically, and select promising candidates. Its main limitation
is that syntactic diversity does not necessarily imply economic diversity. A GP population may contain many distinct formulas that produce similar
exposures or return streams. In addition, noisy empirical fitness can
preserve formulas that fit historical artifacts rather than stable
predictive relationships.

\subsection{RL-based Sequential Formula Generation}
\label{rl-based-sequential-formula-generation}

RL-based methods replace random symbolic variation with a learned
construction policy. Instead of mutating a complete expression, the system builds a formula step by step, selecting fields, operators, windows, or grammar actions conditioned on the partial expression
\cite{yu_et_al_2023,mnih_et_al_2015,bellemare_et_al_2017,dabney_et_al_2018}.
AlphaGen applies proximal policy optimization to expression construction
and introduces a synergistic reward that measures the incremental
contribution of a candidate factor to an existing pool
\cite{yu_et_al_2023}. Alpha$^2$ extends this direction through
trajectory-level reward shaping \cite{xu_alpha2_2024}, while AlphaQCM
uses distributional RL to model return-related uncertainty
\cite{zhu_alphaqcm_2025}.

Within the framework, RL primarily strengthens the variation component. Formula construction is no longer a blind mutation process; it becomes a sequential decision process optimized using feedback from previous evaluations. RL can also provide limited memory through trajectories, replay buffers, or policy histories. Selection is typically handled through reward optimization or candidate-pool update rules. The fitness signal, however, remains the main bottleneck. When the reward is derived from fixed historical data, the policy may learn to exploit sample-specific artifacts. Sparse rewards, policy collapse, and nonstationary reward distributions remain important obstacles.

\subsection{Pool-Aware and Combination-Oriented Mining}
\label{pool-aware-and-combination-oriented-mining}

A formula may perform well in isolation but add little to a portfolio if it is redundant with existing signals. Pool-aware methods therefore link alpha discovery more directly to factor combination and downstream deployment. AlphaForge couples factor generation with dynamic combination by first constructing a diverse candidate pool and then learning to select or weight factors at the portfolio level \cite{shi_alphaforge_2025}.

This family primarily strengthens the selection component. The retained pool is not a passive list of high-scoring formulas; it is evaluated according to its incremental contribution to a combined strategy. This view moves alpha
discovery closer to practical deployment because a useful factor should
add residual value to the existing pool. The unresolved issue lies in the
quality of the first stage. If candidate generation is still guided by
noisy raw IC, the pool may already be populated with overfitted formulas
before dynamic combination begins.

\subsection{Diversity-oriented and Distributional Alpha Search}
\label{diversity-oriented-and-distributional-alpha-search}

Diversity-oriented methods respond to the tendency of GP and RL searches
to converge to a small number of high-scoring modes. AlphaSAGE applies
GFlowNets to formulaic alpha discovery, learning to sample expression
families in proportion to reward rather than returning only a single
optimization trajectory \cite{chen_alphasage_2025}. Related
quality-diversity perspectives emphasize archives that preserve multiple
high-quality behavioral niches
\cite{lehman_and_stanley_2011a,pugh_et_al_2016,flageat_and_cully_2024}.

The main contribution of this family is to make diversity part of the
search process rather than only a post-discovery filter. In the
six-component framework, variation becomes distributional, and selection
is designed to preserve multiple promising regions. The limitation lies
in the distinction between structural diversity and economic diversity.
Two formulas may differ in syntax, tree shape, or construction trajectory
but still capture the same market exposure. Therefore, diversity should
be assessed behaviorally, for example, through factor correlation,
exposure similarity, return-stream similarity, and marginal contribution
to a factor pool.

\begin{table*}[!t]
\caption{Method families mapped to the six-component framework.}
\label{tab:family-framework}
\centering
\scriptsize
\setlength{\tabcolsep}{2.3pt}
\renewcommand{\arraystretch}{1.48}
\newcolumntype{Y}{>{\raggedright\arraybackslash}X}

\begin{tabularx}{\textwidth}{@{}
>{\raggedright\arraybackslash}p{0.13\textwidth}
>{\raggedright\arraybackslash}p{0.145\textwidth}
>{\centering\arraybackslash}p{0.028\textwidth}
>{\centering\arraybackslash}p{0.028\textwidth}
>{\centering\arraybackslash}p{0.030\textwidth}
>{\centering\arraybackslash}p{0.030\textwidth}
>{\centering\arraybackslash}p{0.035\textwidth}
>{\centering\arraybackslash}p{0.028\textwidth}
Y
>{\raggedright\arraybackslash}p{0.155\textwidth}@{}}
\toprule
Family & Representative Methods & R & V & F$^{\dagger}$ & S & M & A &
Main Contribution & Main Limitation \\
\midrule

Human-guided formula libraries &
Alpha101, Alpha191, and Qlib baselines &
\covStrong & \covWeak & \covWeak & \covWeak & \covWeak & \covWeak &
Interpretable symbolic priors and reusable factor templates &
Limited scalability and manual search bias \\

GP- and EA-based symbolic search &
AutoAlpha and AlphaEvolve &
\covModerate & \covStrong & \covWeak & \covModerate & \covWeak & \covWeak &
Population-based search over symbolic expressions &
Fitness noise, bloat, semantic redundancy, and limited validation reliability \\

RL-based sequential generation &
AlphaGen, Alpha$^{2}$, and AlphaQCM &
\covModerate & \covStrong & \covWeak & \covModerate & \covModerate & \covWeak &
Policy-based sequential formula construction &
Sparse rewards, policy collapse, and dependence on fixed historical environments \\

Pool-aware and combination-oriented mining &
AlphaForge and dynamic combination methods &
\covModerate & \covModerate & \covWeak & \covStrong & \covModerate & \covWeak &
Integration of alpha discovery with factor-pool contribution &
Candidate pools may inherit first-stage overfitting and validation bias \\

Diversity-oriented and graph-based search &
AlphaSAGE and AlphaPROBE &
\covModerate & \covStrong & \covWeak & \covStrong & \covModerate & \covWeak &
Distributional search and structured factor relations &
Structural diversity may not imply economic diversity or out-of-sample robustness \\

MCTS and grammar-constrained search &
RiskMiner and alphaCFG &
\covModerate & \covStrong & \covWeak & \covStrong & \covModerate & \covWeak &
Budget allocation through tree expansion and confidence-guided search &
High rollout cost, reward-stationarity assumptions, and validation leakage risk \\

LLM-guided formula generation &
Alpha-GPT and FAMA &
\covStrong & \covStrong & \covWeak & \covWeak & \covWeak & \covWeak &
Reasoning-driven proposal, formula repair, and semantic generation &
Prompt sensitivity, hallucination, and dependence on external validation \\

Agentic and self-evolving discovery &
AlphaAgent, FactorMiner, RD-Agent, and QuantEvolver &
\covStrong & \covStrong & \covModerate & \covModerate & \covStrong & \covModerate &
Tool use, structured memory, coordination, reflection, and early closed-loop self-improvement &
Reproducibility risks, unvalidated memory updates, meta-overfitting, and unstable feedback \\

\bottomrule
\multicolumn{10}{@{}p{\textwidth}@{}}{\footnotesize
\emph{Note:} This table is a component coverage map rather than a performance comparison.
R = representation, V = variation, F = fitness evaluation, S = selection, M = memory, and A = adaptation.} \\
\multicolumn{10}{@{}p{\textwidth}@{}}{\footnotesize
$^{\dagger}$The F column evaluates the explicit treatment of fitness reliability and validation design, rather than the mere use of a reward, IC, RankIC, or backtesting score.} \\
\multicolumn{10}{@{}p{\textwidth}@{}}{\footnotesize
\covStrong{} = strong explicit modeling, \covModerate{} = moderate or partial implementation, and
\covWeak{} = weak, implicit, or absent treatment.}
\end{tabularx}
\end{table*}

\subsection{Graph-Based Retrieval and Structured Factor Pools}
\label{graph-based-retrieval-and-structured-factor-pools}

Graph-based methods treat discovered factors as related objects rather
than independent expressions. AlphaPROBE represents the factor pool as a directed acyclic graph, where dependency or similarity relations among expressions are used to guide retrieval during search \cite{guo_alphaprobe_2026}. This representation allows the discovery system to reuse structural relationships, identify neighborhoods of related factors, and avoid treating the archive as a flat list.

In the framework, graph-based methods mainly enrich representation and memory.
Representation no longer describes only individual formulas, but also captures relations among factors. Memory becomes a structured repository that can be queried and updated according to topology. The main benefit is improved search efficiency and more explicit management of factor relationships. The trade-off is a stronger dependence on archive quality. If the graph is built from noisy or redundant factors, subsequent retrieval may steer the search toward weak or crowded regions.

\subsection{MCTS and Grammar-Constrained Search}
\label{mcts-and-grammar-constrained-tree-search}

MCTS-based methods bring explicit exploration control to symbolic formula search. RiskMiner uses risk-seeking MCTS to pursue trajectories
with high reward variance, rather than only high mean reward
\cite{ren_riskminer_2024}. Grammar-constrained tree search methods, such as
alphaCFG and related work, combine upper-confidence search with formal
expression constraints to improve validity and budget allocation
\cite{shi_alpha_jungle_2026,yang_alphacfg_2026}. These methods build on the success of MCTS in large discrete search spaces
\cite{silver_et_al_2016,coulom_2006}.

Within the six-component framework, MCTS mainly strengthens variation, selection, and local memory. New candidates are generated by expanding partial
expressions, while node selection is guided by visit counts, reward
estimates, and upper-confidence bounds. The search tree also stores local experience about explored states. This gives the system a structured way to allocate the evaluation budget, but it does not remove the cost of rollouts or the noise in empirical rewards. When the reward distribution changes across market regimes, a tree policy calibrated on past evaluations may no longer be reliable.

\subsection{LLM-guided Formula Generation}
\label{llm-guided-alpha-generation}

LLM-guided methods introduce language priors and reasoning-based proposal
mechanisms into alpha generation. Alpha-GPT uses chain-of-thought
prompting to guide intermediate reasoning before producing factor
expressions \cite{wang_alpha_gpt_2025}. FAMA introduces a
neural-symbolic factor mining agent built around the Chain of Symbol
paradigm \cite{li_fama_2024}. These methods are related to broader studies on LLM reasoning, tool use, and planning
\cite{brown_et_al_2020,wei_et_al_2022,yao_et_al_2023a,schick_et_al_2023,yao_et_al_2023b}.

The primary effect of LLMs is on representation and variation. Candidate
formulas can be generated from natural-language hypotheses, repaired
according to symbolic constraints, or refined through reasoning. This can
make the search process more interpretable and inject useful domain
priors. However, LLM-guided generation also introduces new failure modes,
including prompt sensitivity, hallucinated rationales, model-version
dependence, and nondeterministic decoding. As a result, LLM outputs
should be treated as candidate proposals rather than validated alphas.

\subsection{Agentic and Self-Evolving Discovery}
\label{agentic-and-self-evolving-discovery-systems}

Agent-based systems extend alpha generation into a broader research
workflow. AlphaAgent uses specialized agents coordinated through
structured debate \cite{tang_alphaagent_2025}. RD-Agent organizes research,
development, and feedback phases with a scheduler for effort allocation
\cite{li_rd_agent_quant_2025}. FactorMiner addresses correlation
crowding through reusable skills, structured experience memory, and a
retrieve--generate--evaluate--distill loop \cite{wang_factorminer_2026}.
These systems are connected to broader studies on multiagent research
automation and autonomous scientific discovery
\cite{wu_autogen_2023,boiko_et_al_2023,zheng_et_al_2025}.

Compared with earlier generators, agent-based systems invest more heavily
in memory, tool use, coordination, and iterative refinement. They come
closest to a closed autonomous discovery loop because they can generate
hypotheses, call tools, evaluate results, store experience, and revise
future actions. Their main limitation is reliability. A system that can
generate and test more hypotheses without stronger validation may amplify
false discoveries more rapidly. Reasoning-driven variation, structured
memory, and agent-mediated selection are useful only when the fitness
signal, memory-update rule, and adaptation mechanism are themselves
validated under changing market conditions.

\subsection{Comparative Synthesis}
\label{comparative-synthesis}

The reviewed method families differ in how they allocate effort across
the six components. Table~\ref{tab:family-framework} summarizes this
family-level comparison. The table should be interpreted as a component
coverage map rather than a performance comparison. Strong indicates that
the component is explicitly modeled and discussed within the method
family. Moderate indicates that the component is partially implemented or
indirectly supported. Weak indicates that the component is absent,
implicit, or not independently analyzed in the reviewed evidence.

The main pattern in Table~\ref{tab:family-framework} is structural
asymmetry. Human-guided and GP-based methods established the symbolic
representation and population-search template. RL methods improved
sequential generation. GFlowNet-based, graph-based, and MCTS-based
methods strengthened diversity, structure, and exploration control.
LLM-guided and agentic systems added language priors, tool use,
memory, and workflow automation. Across this trajectory, however,
representation and variation have advanced faster than fitness
reliability, validated memory, and active adaptation.

This synthesis leads to a conservative conclusion. The field has moved
from manual formula libraries toward increasingly autonomous evolutionary
discovery systems, but increasing loop closure does not guarantee
reliable alpha discovery. If the empirical fitness signal remains noisy
and the memory-update rule is not validated, a more autonomous system may
learn to exploit historical artifacts more efficiently. This observation
motivates the evaluation protocol and roadmap discussed in
Section~\ref{evaluation-protocol-and-future-roadmap}.

\begin{table*}[!t]
\caption{Autonomy-oriented evaluation matrix for formulaic alpha discovery.}
\label{tab:evaluation-protocol}
\centering
\scriptsize
\setlength{\tabcolsep}{2.1pt}
\renewcommand{\arraystretch}{1.42}
\begin{tabularx}{\textwidth}{@{}
>{\raggedright\arraybackslash}p{0.125\textwidth}
>{\raggedright\arraybackslash}p{0.125\textwidth}
>{\raggedright\arraybackslash}X
>{\raggedright\arraybackslash}p{0.225\textwidth}
>{\raggedright\arraybackslash}p{0.185\textwidth}
>{\raggedright\arraybackslash}p{0.105\textwidth}@{}}
\toprule
Evaluation Dimension &
Linked Components &
Core Question &
Recommended Evidence &
Minimum Reporting Items &
References \\
\midrule

Search efficiency &
Variation, selection &
Does the system explore the formula space efficiently under a fixed and
disclosed search budget? &
Candidate count, valid-expression ratio, evaluation budget, convergence
curve, duplicate rate, and search time &
Budget, hardware, stopping rule, operator set, number of evaluated
candidates, and random seeds &
\cite{jones_et_al_1998,forrester_et_al_2008} \\
\addlinespace[1pt]

Fitness reliability &
Fitness evaluation, selection &
Is the empirical fitness signal robust to sampling noise, leakage,
selection bias, and repeated testing? &
Multiple-testing correction, resampling validation, confidence intervals,
purged validation, and stability across seeds &
Fitness definition, universe construction, validation split, leakage
controls, survivorship-bias control, correction method, seed sensitivity,
and confidence estimates &
\cite{lopez_de_prado_2018,white_2000,bailey_and_lopez_de_prado_2014,bailey_et_al_2017} \\
\addlinespace[1pt]

Residual alpha quality &
Fitness evaluation, selection &
Does the factor retain incremental predictive value after controlling for
known risks and existing factors? &
Risk-neutral IC, residual return, exposure analysis, and incremental pool
contribution &
Risk model, neutralization fields, benchmark factors, residual metric, and
pool baseline &
\cite{fama_and_french_1993,carhart_1997,fama_and_french_2015,grinold_and_kahn_2000} \\
\addlinespace[1pt]

Economic diversity &
Representation, selection, memory &
Do discovered factors represent complementary economic mechanisms rather
than redundant symbolic forms? &
Factor correlation, return-stream similarity, exposure distance,
behavioral descriptors, and marginal contribution &
Correlation threshold, diversity measure, exposure report, pool overlap,
and redundancy filter &
\cite{lehman_and_stanley_2011a,pugh_et_al_2016,tsay_2010} \\
\addlinespace[1pt]

Tradability &
Fitness evaluation, selection &
Can the discovered factor survive realistic implementation frictions and
still remain useful for candidate selection? &
Turnover, liquidity exposure, transaction-cost analysis, slippage analysis,
capacity tests, drawdown, and holding-period sensitivity &
Cost model, slippage model, turnover, liquidity filter, capacity assumption,
rebalancing rule, and net-return metric &
\cite{almgren_and_chriss_2001,frazzini_et_al_2019} \\
\addlinespace[1pt]

Evolutionary autonomy &
Memory, variation, selection &
Does the system reuse experience in a way that improves subsequent
validated search? &
Memory ablation, archive validation, retrieval accuracy, tracking of
failed cases, and search improvement across episodes &
Memory structure, update rule, retrieval rule, ablation setting, and
validation rule for stored entries &
\cite{schmidhuber_1987,shinn_et_al_2023,madaan_et_al_2023} \\
\addlinespace[1pt]

Nonstationary robustness &
Adaptation, memory, fitness evaluation &
Does the system respond to changing market regimes and moving fitness
landscapes? &
Walk-forward tests, regime-split analysis, decay analysis, rolling
validation, and adaptation ablation &
Regime definition, adaptation rule, validation windows, retraining schedule,
rolling protocol, and performance by regime &
\cite{branke_2001,jin_and_branke_2005,farina_et_al_2004,avellaneda_and_lee_2010} \\
\addlinespace[1pt]

Reproducibility &
All components &
Can the complete discovery process be independently verified? &
Open code, fixed seeds, data-construction details, search logs, full
configuration, and rejected-candidate statistics &
Code, data version, preprocessing procedures, operator library, seeds,
search budget, data splits, and evaluation scripts &
\cite{pineau_et_al_2021,peng_2011,stodden_et_al_2016,wilson_et_al_2014} \\

\bottomrule
\end{tabularx}
\end{table*}

\section{Evaluation Protocol and Future Roadmap}
\label{evaluation-protocol-and-future-roadmap}

The taxonomy in
Section~\ref{taxonomy-of-automated-alpha-discovery-methods} shows that
automated formulaic alpha discovery has become increasingly autonomous,
but has not necessarily become more reliable. Recent systems can generate
more candidate expressions, coordinate a wider range of tools, store more
search experience, and revise more intermediate decisions. However, their
evaluation still often relies on noisy backtesting statistics,
inconsistent reporting, and limited validation under changing market
conditions. This section converts this gap into an autonomy-oriented
evaluation protocol and identifies the main roadmap issues for reliable
alpha discovery.

\subsection{An Autonomy-Oriented Evaluation Protocol}
\label{an-autonomy-oriented-evaluation-protocol}

A candidate alpha is usually evaluated by applying its symbolic expression
across instruments and time periods and then measuring whether the
resulting scores are associated with future returns. Common metrics
include predictive association, temporal stability, risk-adjusted signal
quality, implementation feasibility, and generalization beyond the
original backtesting setting \cite{grinold_and_kahn_2000}. These metrics
are necessary, but they are not sufficient for autonomous discovery
systems. An autonomous system is not only a predictor; it is also a search
process that repeatedly generates, evaluates, selects, stores, and adapts
candidate formulas. Therefore, evaluation should assess both the quality
of the discovered factors and the reliability of the discovery process
that produced them.

The difficulty is that financial fitness is not a fixed objective. It is
an empirical estimate of future predictive utility. It is affected by
sampling noise, cross-sectional dependence, market regimes, repeated
testing, transaction costs, and implementation assumptions. Two formulas
with similar latent value can receive different observed information
coefficient (IC) or rank information coefficient (RankIC) values. A
formula selected in one market regime may also decay in another. As a
result, evaluating only the best reported IC or long--short return is
insufficient for comparing automated formulaic alpha discovery systems.

Current benchmarks only partially address this issue. Common Qlib-style
settings improve accessibility and reproducibility, but repeated
optimization against the same market, frequency, split, operator library,
and cost assumptions can create collective overfitting
\cite{recht_et_al_2019}. Statistical significance testing, seed
sensitivity, search-budget disclosure, transaction costs, neutralization
protocols, and regime robustness are still reported inconsistently
\cite{chan_et_al_2020,agarwal_et_al_2021,pineau_et_al_2021}. This is
especially problematic for LLM-guided and agentic systems, because
their larger search scope and richer memory can amplify false discoveries
when evaluation is weak.

Building on the six-component framework, this article proposes an
autonomy-oriented evaluation protocol with eight dimensions.

\begin{enumerate}
\item \emph{Search efficiency.} This dimension evaluates whether the
system explores the symbolic search space efficiently under a disclosed
budget \cite{forrester_and_keane_2009,hutter_et_al_2011}.

\item \emph{Fitness reliability.} This dimension evaluates whether the
fitness signal is statistically reliable under noise, leakage, repeated
testing, and validation uncertainty.

\item \emph{Residual alpha quality.} This dimension evaluates whether a
factor provides incremental predictive power beyond known risk exposures,
style factors, and existing alpha libraries.

\item \emph{Economic diversity.} This dimension evaluates whether
discovered factors capture genuinely complementary return sources rather
than redundant symbolic variants.

\item \emph{Tradability.} This dimension evaluates whether turnover,
capacity, liquidity, and transaction-cost constraints are incorporated
into the assessment.

\item \emph{Evolutionary autonomy.} This dimension evaluates whether
memory and experience reuse improve subsequent validated search rather
than reinforcing spurious discoveries.

\item \emph{Nonstationary robustness.} This dimension evaluates whether
the discovery system remains effective under regime shifts and changing
fitness landscapes.

\item \emph{Reproducibility.} This dimension evaluates whether the
discovery process can be independently verified through disclosed data,
code, seeds, configurations, splits, and search logs.
\end{enumerate}

Table~\ref{tab:evaluation-protocol} summarizes the proposed evaluation
matrix. The table links each dimension to the six-component framework,
states the core evaluation question, and lists the minimum reporting items
needed for reproducible comparison. The matrix is not intended to replace
domain-specific metrics. Instead, it defines the minimum evidence needed
to judge whether an autonomous discovery system is reliable.

Table~\ref{tab:evaluation-protocol} also suggests several broader implications. Fitness reliability is the most immediate bottleneck because every other component follows the fitness signal it receives. A stronger
generator, a larger memory, or a more complex agent workflow cannot
compensate for an unreliable empirical objective. Residual alpha quality and economic diversity are also necessary for practical deployment. A
factor with high raw IC but strong overlap with known exposures or
existing factors may provide little incremental value. Third,
evolutionary autonomy and reproducibility should be evaluated as system
properties, not as informal descriptions. A memory module is useful only
if it improves future validated discovery, and an autonomous workflow is
scientific only if its search process can be independently reconstructed.

\subsection{Reliable Fitness under Noisy and Nonstationary Markets}
\label{reliable-fitness-under-noisy-and-nonstationary-markets}

The first research priority is reliable fitness under noisy and changing
market conditions. Most reviewed methods still optimize raw IC, RankIC, or
closely related backtesting statistics, although candidate scores are
estimated from finite samples, correlated instruments, changing regimes,
and large implicit hypothesis searches. When the noise-to-signal ratio is
high, stronger variation operators may discover spurious patterns more
quickly rather than uncover more reliable alphas
\cite{beyer_and_schwefel_2002,jin_and_branke_2005}.

A more reliable fitness protocol should address three issues within the
search loop. First, repeated testing should be reflected in selection
thresholds as the number of evaluated candidates grows. The deflated
Sharpe ratio, superior predictive ability tests, and the probability of
backtest overfitting provide relevant statistical tools
\cite{ledoit_and_wolf_2008,bailey_and_lopez_de_prado_2014,white_2000,bailey_et_al_2017}.
Second, resampling should respect temporal dependence through purged
validation, walk-forward testing, and cross-market validation
\cite{lopez_de_prado_2018}. Third, fitness should be conditioned on
regime information when market states differ structurally. Factor decay
and macroeconomic shifts make the optimum time-dependent rather than
fixed \cite{avellaneda_and_lee_2010}. Averaging performance across
incompatible regimes may hide precisely the failures that autonomous
systems need to detect.

The practical target is a fitness signal that combines control of
repeated testing, resampling-based validation, and regime-conditioned
performance. Without such a signal, increasing autonomy mainly increases
the speed at which systems fit historical noise.

\subsection{From Syntactic Diversity to Economic Diversity}
\label{from-syntactic-diversity-to-economic-diversity}

Diversity has become an important objective in recent alpha discovery
systems, but many implementations still measure it through formula
structure or historical correlation. This is insufficient. Two formulas
may have different syntax, different expression trees, or even low
average correlation while exploiting the same economic return source.

Useful diversity has three levels. Syntactic diversity measures
differences in tokens, trees, or grammar derivations. Semantic diversity
requires behavioral differences in factor outputs, exposure profiles, or
response surfaces. Economic diversity asks whether factors earn returns
from different mechanisms, such as information diffusion, liquidity
provision, behavioral biases, institutional frictions, or risk transfer.
The last level is the most relevant to portfolio construction, but it is also the hardest to validate.

Low historical correlation is particularly unreliable during periods of market stress \cite{tsay_2010}. The correlation red-sea problem described by FactorMiner captures this issue: large factor pools may appear diverse under structural measures but become highly redundant when used in portfolios \cite{wang_factorminer_2026}. A more reliable diversity objective should combine behavioral descriptors, risk exposures, turnover profiles, capacity constraints, and regime-conditioned comovement, rather than relying only on formula distance or token novelty.

\subsection{Market-Logic Grounding and Interpretability}
\label{market-logic-grounding-and-interpretability}

LLM-guided methods make it easier to attach natural-language rationales
to generated formulas, but a coherent rationale is not evidence that a
factor is economically valid. Market logic is useful only when it acts as
a testable constraint on generation and evaluation.

Market-logic libraries provide one possible mechanism. They can store
reusable patterns grounded in market microstructure, investor behavior,
and institutional frictions, allowing factor generation to begin from an
explicit hypothesis rather than a purely correlational search
\cite{weng_alphalogics_2026}. LLM-based agent systems can then translate these
hypotheses into candidate formulas and expose the rationales for expert
review \cite{tang_alphaagent_2025}. The risk is narrative overfitting. An LLM
can generate a plausible explanation for a spurious correlation, and
narrative plausibility may make the factor appear more credible than the
supporting evidence warrants.

Future systems should therefore separate hypothesis generation from
hypothesis testing. Agents may propose market-logic-grounded candidates,
but fitness evaluation should remain independent, statistically rigorous,
and insensitive to narrative appeal.

\subsection{Multiobjective Evolutionary Alpha Discovery}
\label{multiobjective-evolutionary-alpha-discovery}

Raw IC alone is an insufficient objective for autonomous discovery because
the desirable properties of a factor are not aligned along a single axis.
Predictive power, stability, turnover, capacity, complexity, novelty, and
residual value can conflict with one another. A factor with slightly lower
IC may be preferable if it is more stable, cheaper to trade, or less
redundant.

Alpha discovery is therefore better formulated through the multiobjective
fitness vector defined in Eq.~\eqref{eq:multiobjective-alpha}.
In evaluation, this vector can be operationalized by IC or RankIC for
predictive quality, ICIR and regime-wise validation for stability, factor
correlation or residual contribution for diversity and novelty, turnover
and transaction-cost-adjusted return for tradability, expression complexity
for simplicity and interpretability, and code, data, seed, and reporting
completeness for reproducibility \cite{deb_and_jain_2014,farina_et_al_2004,emmerich_and_deutz_2018}.

This formulation connects alpha discovery to established EC tools. Pareto
optimization can represent trade-offs between predictive power and
tradability \cite{deb_et_al_2002,zhang_and_li_2007}. Quality-diversity
methods can maintain archives across behavioral niches
\cite{mouret_and_clune_2015}. Constrained evolutionary optimization can
enforce turnover, capacity, leverage, or risk-exposure constraints during
search rather than after selection. These tools are not optional
refinements. They are closer to the actual decision problem faced by
deployable alpha systems.

\subsection{Surrogate-Assisted and Costly Fitness Evaluation}
\label{surrogate-assisted-and-costly-fitness-evaluation}

Fitness evaluation is sufficiently costly to constrain search, but
inexpensive proxy evaluation can introduce its own failure modes.
Comprehensive backtesting across many instruments and time periods constrains population size, resampling depth, and search budget. Surrogate-assisted EC addresses this constraint through learned evaluators, proxy fitness functions, Bayesian optimization, and multifidelity scheduling
\cite{jin_and_branke_2005,shahriari_et_al_2016,brochu_et_al_2010,knowles_2006}.

The main risk is surrogate bias. A proxy model that systematically
misestimates novel expression structures may steer search toward a proxy
optimum rather than true economic utility. It may also prune useful candidates too early or allocate costly backtests to false positives. The issue is not only whether a surrogate is fast, but also whether it can quantify its own uncertainty.

Future systems should use uncertainty-aware surrogates to decide which
candidates deserve full evaluation. Candidates with promising,
out-of-distribution, or high-uncertainty proxy estimates should receive
more costly validation. Graph memory and factor retrieval may also
reduce cost by reusing information from structurally related formulas.
The key requirement is disciplined allocation of backtests, not merely a
faster approximation of raw IC.

\subsection{Validated Memory, Reflection, and Self-Evolution}
\label{validated-memory-reflection-and-self-evolution}

Memory is the component that most clearly separates a one-shot generator
from a self-improving discovery system. It also creates a direct path for
error accumulation. A memory archive that stores overfitted factors,
misleading rationales, or lucky trajectories will guide subsequent search
toward the same failures.

Current agentic systems have begun to introduce structured memory.
FactorMiner stores reusable skills and factor experience, while RD-Agent
records research and development feedback across iterations
\cite{wang_factorminer_2026,li_rd_agent_quant_2025}. Such memory can include
successful factors, failed trials, tool-use traces, search trajectories,
and market-logic patterns. Its value, however, depends on validation.

A reliable memory system should periodically reevaluate archived factors
on new data, purge entries that fail statistical tests, and weight
retrieval according to the reliability of the original evaluation. The
same principle applies to reflection. Prompt-level self-critique is not
equivalent to policy-level self-evolution. QuantEvolver-style repositories
represent early steps toward longitudinal policy updates
\cite{zhang_quant_evolver_2026}, but meta-overfitting remains unresolved. A
useful memory system should be judged by whether it improves future
validated discovery, not by how many trajectories or rationales it stores.

\subsection{Human-in-the-Loop Autonomous Quantitative Research}
\label{human-in-the-loop-autonomous-quant-research}

As autonomous systems generate and test more candidates, the role of human researchers shifts from writing individual formulas to governing the search process. This shift is necessary because fully automated systems can
scale false discoveries as easily as useful exploration.

Human oversight should define the scientific and risk-management boundaries within which the search is conducted. Researchers can specify which market mechanisms are plausible, which exposures are unacceptable, how statistical significance is assessed, and what level of evidence is required before a factor enters production. Agents can then search, evaluate, and refine candidates within these boundaries.

Such oversight does not weaken autonomy. It makes autonomy auditable. Agents
are well-suited to combinatorial search and systematic evaluation, while
human experts remain essential for causal judgment, regime
interpretation, and risk governance. Future systems need interfaces that
translate natural-language risk constraints into evolutionary penalties,
expose uncertainty and failure modes, and support human review before
deployment.

\subsection{Reproducible Benchmarks and Reporting Standards}
\label{reproducible-benchmarks-and-reporting-standards}

Without shared reporting standards, performance differences across alpha
discovery systems remain difficult to interpret. Datasets, operators,
search budgets, seeds, evaluation windows, and cost assumptions vary
widely across studies, making cross-method comparison unreliable.

Autonomous systems make this problem more severe because they test more
hypotheses and introduce additional stochasticity through search
policies, LLM decoding, agent coordination, and memory updates
\cite{peng_2011,stodden_et_al_2016,wilson_et_al_2014}. Future benchmarks
should disclose dataset construction, preprocessing rules, operator
libraries, search budgets, random seeds, fitness definitions, selection
rules, memory configurations, adaptation rules, transaction costs,
neutralization protocols, out-of-sample splits, out-of-distribution
tests, and code availability. These requirements are summarized in
Table~\ref{tab:evaluation-protocol}.

For autonomous systems, reproducibility must include the search process
itself. Reports should include candidate counts, LLM calls, stopping rules, agent coordination mechanisms, rejected candidates, and all evaluation filters. Without such information, it is difficult to distinguish genuine algorithmic progress from undisclosed search budgets, lucky random seeds, or uncorrected repeated testing.

\subsection{Toward Scientific Automated Formulaic Alpha Discovery}
\label{toward-scientific-autonomous-alpha-discovery}

The roadmap developed above points to a clear conclusion: automated formulaic alpha discovery cannot advance through search sophistication alone. The field needs systems that search efficiently, evaluate reliably, adapt to changing regimes, and support reproducible discovery.

These priorities are interconnected. Reliable fitness determines whether search is guided by signal rather than noise. Diversity and tradability determine whether discovered factors add portfolio value. Market logic constrains generation by requiring testable hypotheses. Memory and adaptation determine whether the system improves over time or simply accumulates noise. Human oversight and reporting standards determine whether reported progress can be independently assessed.

Realizing this agenda will require collaboration among EC researchers,
financial econometricians, and quantitative practitioners. The relevant benchmark is therefore not the complexity of the search architecture, but the scientific validity of the discoveries it produces.

\section{Conclusion}
\label{conclusion}

This article has examined automated formulaic alpha discovery from an evolutionary computation perspective. Instead of viewing existing studies as separate algorithmic families, it frames the field as a form of noisy and dynamic symbolic evolutionary optimization. Under this formulation, candidate formulaic alphas are symbolic genotypes, empirical backtesting metrics are noisy fitness signals, and discovery proceeds through a closed loop of representation, variation, fitness evaluation, selection, memory, and adaptation.

The six-component framework shows a structural imbalance in the literature.
Representation and variation have become increasingly sophisticated through
symbolic encodings, learned policies, distributional sampling, tree search,
language priors, and agentic workflows. By contrast, fitness reliability,
validated memory, and regime-aware adaptation remain comparatively
underdeveloped. This imbalance is critical because greater autonomy does not
automatically imply more reliable alpha discovery; without robust validation,
autonomous systems may amplify historical artifacts at greater scale.

Future work should therefore shift from formula proliferation to validated discovery systems that can withstand statistical testing, risk control, transaction-cost adjustment, regime change, and independent verification. The central measure of progress is not the complexity of the search machinery, but the validity and practical value of the alpha factors it discovers.

\ifCLASSOPTIONcaptionsoff
  \newpage
\fi

\end{document}